\documentclass[lettersize,journal]{IEEEtran}
\usepackage{amsmath,amsfonts}
\usepackage{algorithmic}
\usepackage{algorithm}
\usepackage{array}
\usepackage[caption=false,font=normalsize,labelfont=sf,textfont=sf]{subfig}
\usepackage{textcomp}
\usepackage{stfloats}
\usepackage{url}
\usepackage{verbatim}
\usepackage{graphicx}
\usepackage{multirow}
\usepackage{cite}
\usepackage{hyperref}
\usepackage{amssymb}
\usepackage[table,xcdraw]{xcolor}
\usepackage{booktabs}
\begin{document}

\title{MV-CC: Mask Enhanced Video Model for Remote Sensing Change Caption }

\author{Ruixun Liu*, Kaiyu Li*, Jiayi Song*, Dongwei Sun, Xiangyong Cao
\thanks{* Equal contribution. Corresponding author: Xiangyong Cao.}
\thanks{Ruixun Liu is with the School of Automation, Xi’an Jiaotong University, Xi’an 710049, China (email: liuruixun6343@gmail.com)}
\thanks{Kaiyu Li is with the School of Software Engineering, Xi’an Jiaotong University, Xi’an 710049, China (email: likyoo.ai@gmail.com)}
\thanks{Jiayi Song, Dongwei Sun, and Xiangyong Cao are with the School of Computer Science and Technology and Ministry of Education Key Lab For Intelligent Networks and Network Security, Xi’an Jiaotong University, Xi’an 710049, China (email: songyangyifei@gmail.com, sundongwei@outlook.com, caoxiangyong@xjtu.edu.cn)}
}



\maketitle

\begin{abstract}
 Remote sensing image change caption (RSICC) aims to provide natural language descriptions for bi-temporal remote sensing images. Since Change Caption (CC) task requires both spatial and temporal features, previous works follow an encoder-fusion-decoder architecture. They use an image encoder to extract spatial features and the fusion module to integrate spatial features and extract temporal features, which leads to increasingly complex manual design of the fusion module. In this paper, we introduce a novel video model-based paradigm without design of the fusion module and propose a Mask-enhanced Video model for Change Caption (MV-CC). Specifically, we use the off-the-shelf video encoder to simultaneously extract the temporal and spatial features of bi-temporal images. Furthermore, the types of changes in the CC are set based on specific task requirements, and to enable the model to better focus on the regions of interest, we employ masks obtained from the Change Detection (CD) method to explicitly guide the CC model. Experimental results demonstrate that our proposed method can obtain better performance compared with other state-of-the-art RSICC methods. The code is available at \url{https://github.com/liuruixun/MV-CC}.
\end{abstract}


\section{Introduction}
\IEEEPARstart{I}{mage} change captioning is an emerging research area at the intersection of computer vision and natural language processing, focused on providing detailed, descriptive accounts of changes in scenes over time~\cite{robustchangecaptioning,imagecaptiioning}. With advancements in remote sensing and earth observation technologies, the availability of bi-temporal remote sensing imagery has increased significantly. With the extensive accumulation of data, there is increasing interest in applying Change Captioning (CC) to study land cover transformations.

Unlike traditional Change Detection (CD) tasks, Remote Sensing Image Change Captioning (RSICC) goes beyond simply locating differences; it generates natural language descriptions that explain how the content of remote sensing images evolves. This capability is pivotal in various fields~\cite{Zhu2024SemanticCCBR,chg2cap}, including urban planning, environmental monitoring, and disaster management. RSICC's ability to translate complex image data into understandable language greatly enhances decision-making in critical areas.


In recent years, research in the RSICC field mainly focused on addressing two critical challenges:

\begin{enumerate}
    \item \textbf{Capturing temporal features.} Unlike traditional single-image tasks \cite{unetconvolutionalnetworksbiomedical}, CC requires not only extracting spatial features from each image but also capturing the temporal features between pairs of images. Building on models pre-trained on a large number of single images, it is essential to design models that accurately capture temporal features.
    \item \textbf{Incorrect attention to regions.} Even if the model captures temporal signals, its performance can still be affected if it focuses on the wrong or inaccurate change regions due to factors like lighting and seasonal variations. Ensuring that the model focuses on the region of change of interest is crucial for improving the performance of CC.
\end{enumerate}

\begin{figure}[t]
  \centering
   \includegraphics[width=0.9\linewidth]{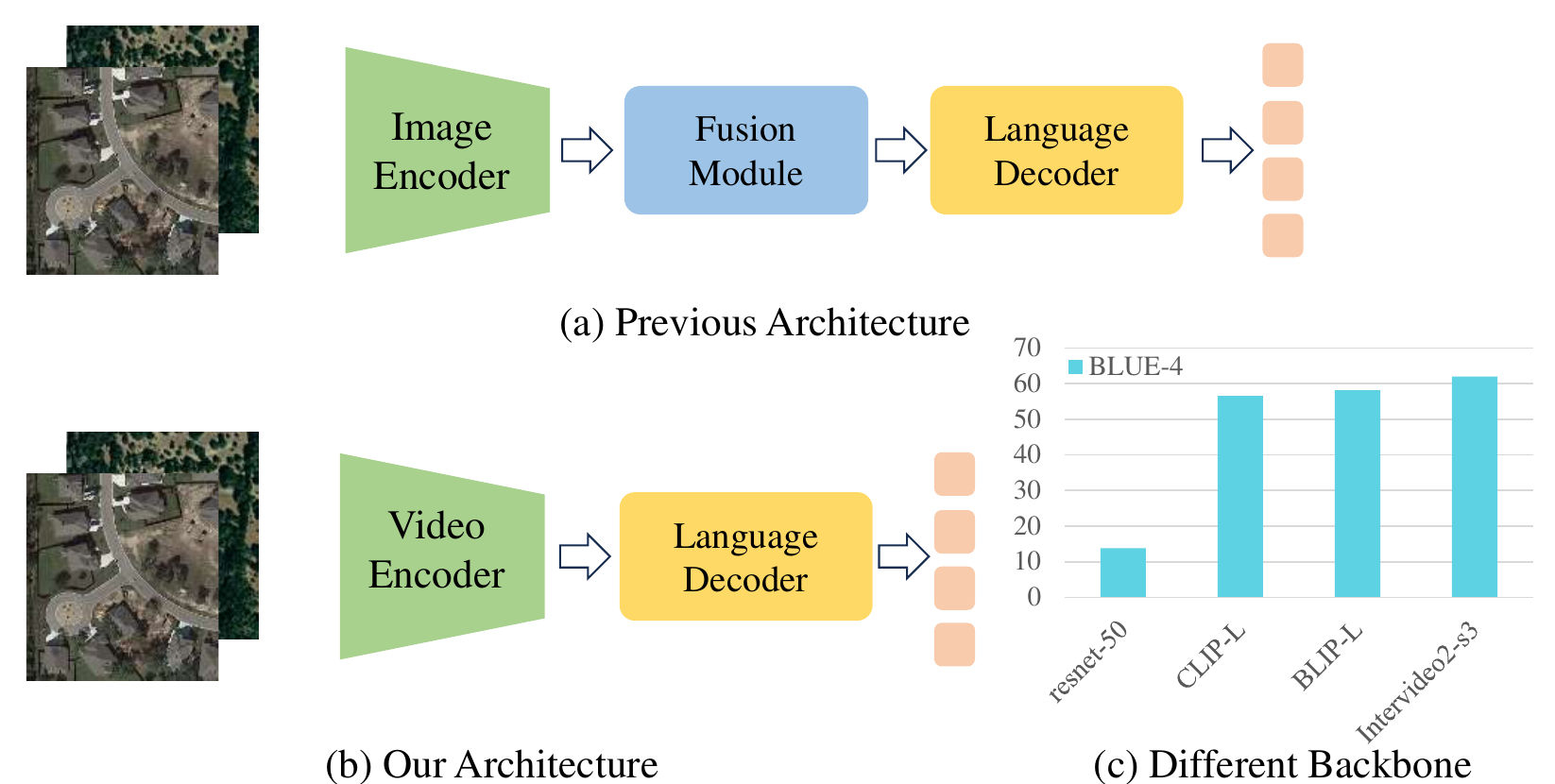}
   \caption{(a) Classical three-stage architecture, including an image encoder, a feature fusion module, and a language decoder. (b) our proposed video-based two-stage architecture. (c) demonstrates the performance of different backbones when the fusion module is omitted.}
   \label{arch}
   \vspace{-4mm}
\end{figure}

Previous work has already achieved significant progress regarding the challenge of capturing temporal features. Classic CC models adopt a three-stage architecture, as shown in Fig. \ref{arch}. This architecture consists of the Siamese spatial image encoder network, the feature fusion module, and the language decoder, where temporal features are captured mainly by the feature fusion module. For instance, Change \textit{et al.} \cite{chg2cap} utilize a Siamese CNN for feature extraction, an Attentive Encoder for feature fusion, and a Transformer decoder for the caption generation.

It is important to note that CC models are developed from CD models. Therefore, CC tasks inherit fusion modules from CD. However, for CD tasks, the CD labels are binary or multi-class dense masks that provide strong supervision, allowing the model to train the fusion module which can effectively capture temporal features. In contrast, for CC tasks, natural language descriptions are more diverse and sparse than masks, leading to weaker supervision signals from the text \cite{imagetextcodecompositiontextsupervisedsemantic}, which hampers the training of the fusion module. Consequently, to enhance the performance of CC models, the design of the fusion module has become increasingly sophisticated. For instance, in \cite{Inter}, a complex symmetric difference Transformer module was designed to enhance the extraction of bi-temporal features. Despite their success, the fusion modules differ significantly in architecture, leaving the optimal design unclear.
\begin{figure}[t]
  \centering
   \includegraphics[width=0.9\linewidth]{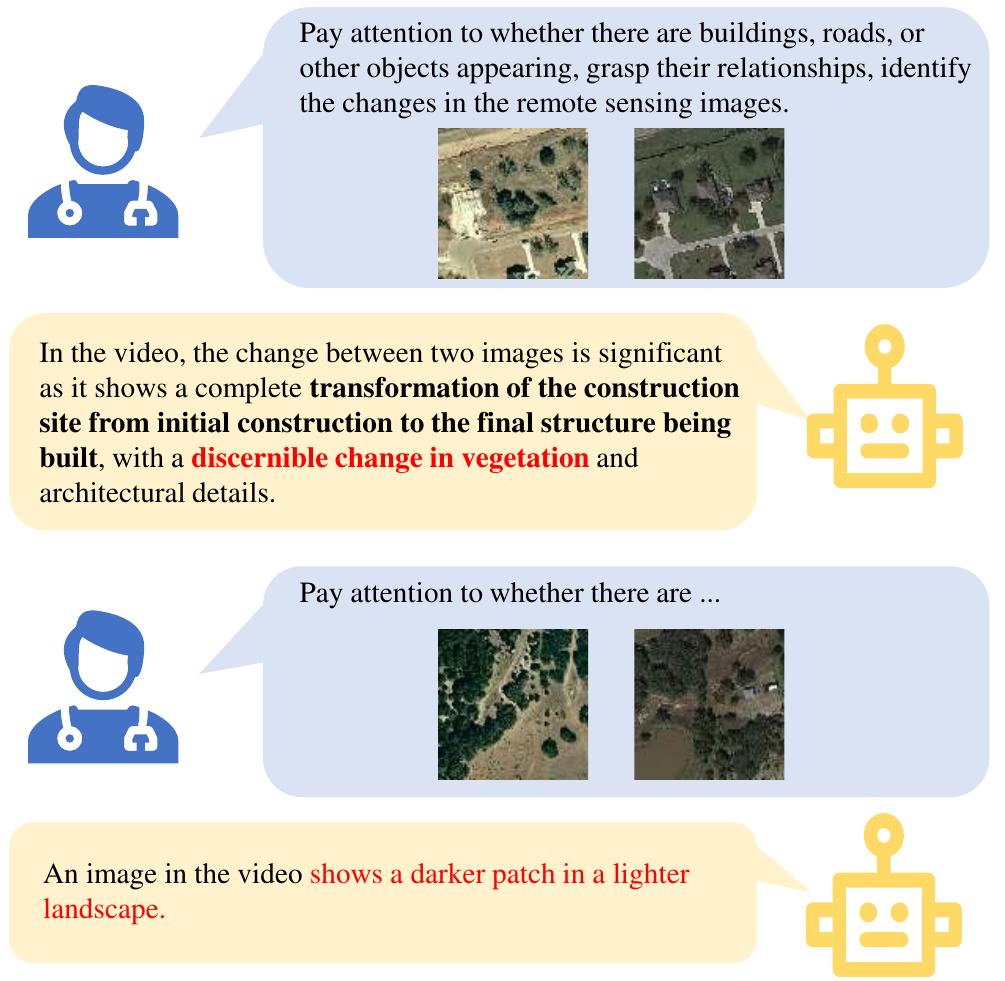}
   \caption{The results of zero-shot inference using InternVideo2. Bold text represents the content of interest, and red text represents the content that is not of interest.}
   \label{interestiong}
   \vspace{-4mm}
\end{figure}

Instead of designing more intricate fusion modules, we aim to explore models that naturally possess the ability to capture temporal signals. Through some exploratory experiments, we found that video models can elegantly conduct spatio-temporal modeling without any elaborate fusion module. As shown in Fig. \ref{arch}(c), we used frozen pre-trained backbone such as ResNet \cite{ResNet} (supervised trained on ImageNet), CLIP \cite{Clip}, and BLIP \cite{Blip}, as well as a video model as feature extractor, feeding their generated features directly into the caption decoder without using any fusion module. The results show that frozen ResNet struggles to capture effective features in remote sensing images. CLIP excels in image-text alignment, while BLIP focuses on generating coherent text from images, leading to improved performance in both models. The video model, which has a similar Transformer structure to CLIP and BLIP while considering temporal modeling during training, significantly outperforms all three. This leads us to believe that video models inherently have a stronger capability to capture temporal signals even in remote sensing images, making them a more advantageous choice for the CC task. This inspired us to discard the complex fusion module and explore the application of video models in CC. 

Since video models are typically trained on natural videos, they often focus on irrelevant changes, such as the density of vegetation and lighting. As shown in Fig. \ref{interestiong}, while the video model can detect differences between two temporal frames, it frequently emphasizes uninteresting change. This phenomenon is observed in video models and has also been described as the semantic noise\cite{Zhu2024SemanticCCBR,10640593} unrelated to the change understanding task in previous CC models. To address this issue of incorrect region attention, several solutions have been proposed in previous studies. For example, Semantic-CC \cite{Zhu2024SemanticCCBR} achieves joint training of the CD and CC models, where the CD mask effectively guides the CC model to focus on relevant change areas. Yang \textit{et al.} \cite{yang2024enhancingperceptionkeychanges} adopted a similar approach. This method provides the model with additional supervisory signals. Li \textit{et al.} \cite{Detection} directly used a pre-trained CD model to generate masks, using these masks as input to guide the CC task.

Inspired by previous work, and noting that change descriptions mainly stem from categories predicted in CD masks, we utilize mature CD techniques to enhance our video model. Unlike previous methods, this work does not use the masks as direct inputs or as additional supervision. Dependent on the inherent captioning capabilities of the video foundation model, our intuition is to filter out irrelevant image context, so that the language model focuses only on the region of change of interest. Therefore, we discard the image tokens corresponding to the unchanged regions in the CD mask, and only a few effective tokens of the changed regions are fed into the language decoder to contribute to the caption generation. Notably, since we filter irrelevant tokens within the feature space, a low-resolution coarse CD mask is sufficient. This weak constraint allows us to utilize various CD models to generate imperfect masks, including semi-supervised and fully supervised methods, and even to directly obtain masks through low-cost manual labeling if possible.


%

Our contributions can be summarized in the following three key aspects:
\begin{enumerate}
    \item Proposing a new paradigm: We introduce a novel paradigm for the CC task, i.e. "video encoder-decoder" architecture. This paradigm utilizes the inherent temporal extraction capabilities of video models and avoids the manual design of complex feature fusion modules. It also has the potential to be directly transferred to multi-temporal CC tasks without an additional design of complex fusion modules.

    \item Explicit mask-based guidance: For the video-based CC model, we propose an explicit guidance method based on CD mask to make it focus only on the change region. Since the guidance only requires low-resolution masks, the coarse masks generated by simple CD models trained on limited supervision signals are sufficient.
    
    \item Our experiments validate the effectiveness and superiority of our proposed video-based models in the CC task compared with other state-of-the-art methods. Additionally, we guided the CC model using various CD models, confirming the effectiveness of CD models in enhancing CC performance.
\end{enumerate}

The rest of this paper is organized as follows: In Section~\ref{sec:work}, we review related work on image change captioning, image change detection in remote sensing, and video models. Section~\ref{sec:methods} presents our proposed MV-CC in detail, showcasing our architecture and framework. Section~\ref{sec:exp}, provides convincing experimental results and Section~\ref{sec:con} concludes our work.
\section{Related Works} \label{sec:work}
\subsection{Image Change Captioning}
\textbf{Change Captioning in natural images.}
Jhamtani \textit{et al.} \cite{jhamtani2018learningdifferencespairssimilar} first introduced the CC task, contributed a dataset, and proposed a Siamese CNN-RNN architecture to capture significant differences between similar images. Part \textit{et al.} \cite{park2019robustchangecaptioning} elevated semantic interpretation to a higher level by introducing a dual dynamic attention model that can differentiate semantic visual information between image pairs, unaffected by variations in lighting and viewpoint. To address viewpoint interference, Shi \textit{et al.} \cite{shi2020findingsideviewpointadaptedmatching} designed a viewpoint-invariant image encoder capable of measuring feature similarity between image pairs in detail. Hosseinzadeh \textit{et al.} \cite{imagecaptiioning} proposed improving the training of change description networks through auxiliary tasks.

\textbf{Change Captioning in Remote Sensing.} To describe the changes in bi-temporal remote sensing images, many studies have recently proposed various methods for RSICC. Chouaf \textit{et al.} \cite{9554419} were pioneers in introducing a CC model that leverages an encoder-decoder framework. Their approach utilized CNN to extract pairs of image features, which were subsequently concatenated and processed by an RNN to produce change descriptions. G. Hoxha \textit{et al.} and Liu \textit{et al.} both contributed new datasets for RSICC\cite{ChangH,LiuDataset}. They also proposed new methods, utilizing CNN architectures for feature extraction. The former used RNN and SVM approaches to generate captions, while the latter employed a dual-branch Transformer encoder to enhance feature discrimination for changes, along with a caption decoder to generate sentences describing the differences. Subsequently, Chang \textit{et al.} proposed the Chg2Cap method \cite{chg2cap}, which employed a Siamese CNN architecture to extract high-level representations of two images, followed by an attentive encoder and caption decoder for generating captions, resulting in significant performance improvements over previous work. To address the weak change recognition issue in RSICC, Liu \textit{et al.} suggested using a pre-trained CD model to generate pseudo-labels to assist the CC task\cite{pixel}, demonstrating the effectiveness of CD task assistance. Following this, many studies have emerged that utilize CD tasks to aid CC \cite{Zhu2024SemanticCCBR,cdchatlargemultimodalmodel,changemindsmultitaskframeworkdetecting}. Zhu \textit{et al.} proposed integrating the CD task \cite{Zhu2024SemanticCCBR} to provide pixel-level semantic guidance for the CC task, using the labels obtained from CD to focus on changed areas of interest in the CC task.

Unlike previous methods, we do not aim to design new modules; instead, we propose to use video encoders to directly extract features from bi-temporal images without an additional fusion module. This is a new paradigm for the CC task.

\subsection{Image Change Detection in Remote Sensing}

\textbf{Semi-Supervised Change Detection.} Semi-supervised CD \cite{li2024semicd_vl, yang2024unimatch} has increased with the development of semi-supervised learning, and it mainly consists of two methods: GAN-based and consistency regularization-based approaches. Peng \textit{et al.} proposed SemiCDNet\cite{SemiCDNet}, which uses a GAN to generate prediction and entropy maps. Two discriminators are employed to maintain the consistency of feature distributions between labeled and unlabeled data. Gedara \textit{et al.} argued that SemiCDNet assumes distribution consistency, limiting its generalization ability. They introduced SemiCD\cite{SemiCD}, which uses consistency regularization to constrain the outputs of unlabeled image pairs to remain consistent under small perturbations. Similarly, Sun \textit{et al.} \cite{SANet} adopted this approach, constructing a Siamese UNet with an attention mechanism, pretraining it with a small amount of labeled data, and filtering pseudo-labels of unlabeled data through confidence thresholds. Mao \textit{et al.} \cite{Mao} applied the Mean Teacher method, training a student model with a small set of labeled data and transferring model parameters to the teacher model via Exponential Moving Average (EMA). The weakly augmented output of the teacher model penalizes the student's strongly augmented predictions.

\textbf{Fully Supervised Change Detection.}
Fully supervised CD has made significant progress in recent years, largely due to the widespread application of CNN and Transformer architectures. Daudt \textit{et al.} were the first to propose an end-to-end Fully Convolutional Network (FCN) architecture, laying the foundation for CD. The subsequent SNUNet-CD\cite{SUNNet} model effectively addressed issues of edge pixel uncertainty and small object omission by combining a densely connected network structure with an Ensemble Channel Attention Module (ECAM), significantly improving target localization accuracy. Meanwhile, the P2V-CD\cite{P2V} framework enhanced detection performance by constructing pseudo-transition videos and decoupling spatiotemporal features, effectively alleviating spatiotemporal coupling problems. In recent years, Transformer architectures have emerged prominently in CD; models like BIT\cite{BiT} and ChangeFormer\cite{ChangeFormer} leverage self-attention mechanisms to capture global dependencies, greatly enhancing detection accuracy and sensitivity to subtle changes.

In this work, we leverage mature CD techniques to guide the CC model in focusing on truly changed regions. By using either semi-supervised or fully supervised methods, we train on existing datasets to obtain change masks for bi-temporal images. These masks are then downsampled and integrated into the CC model through a very simple approach.
\subsection{Video Model}

\textbf{Classic Video Models.} Traditional CNN is an important cornerstone in the field of video understanding. Models like C3D\cite{C3D} and I3D\cite{I3D} effectively capture spatiotemporal features in videos, demonstrating outstanding performance in tasks such as action recognition. Recently, Vision Transformers (ViT)\cite{vit} have also emerged as powerful tools in visual tasks and have been rapidly applied to video understanding. Timesformer\cite{timesformer} employs a mechanism that separates spatial and temporal attention to model spatiotemporal features in videos. Furthermore, ViViT\cite{vivit} expands the application of ViT in the video domain by proposing various architectures and achieving leading performance across multiple tasks.

\textbf{Video Foundation Models.} Recent research has emphasized the importance of learning representations that integrate multimodal information. Noteworthy methods include video-text contrastive learning\cite{vedioclip,Blip,Blip2}, which enhances the model's understanding by aligning video content with textual descriptions. Masked video modeling techniques have also gained attention, VideoMAE\cite{vediomae,vediomaev2} obscures portions of video frames and requires the model to predict the content of the masked areas, thereby improving its spatiotemporal understanding. Additionally, next token prediction\cite{flamingo,generative} serves as a key strategy that enables the model to predict future elements in video or text sequences, which is crucial for understanding temporal information. InternVideo2\cite{Internvideo} integrates mask video modeling, cross-modal contrastive learning, and next-token prediction, demonstrating powerful capabilities in video recognition and captioning. Inspired by this, we treat bi-temporal remote sensing images as an extremely short video to utilize the temporal extraction capability of the video model.
\section{Methods}\label{sec:methods}
\subsection{Architecture}
\begin{figure*}[t]
  \centering
   \includegraphics[width=0.9\linewidth]{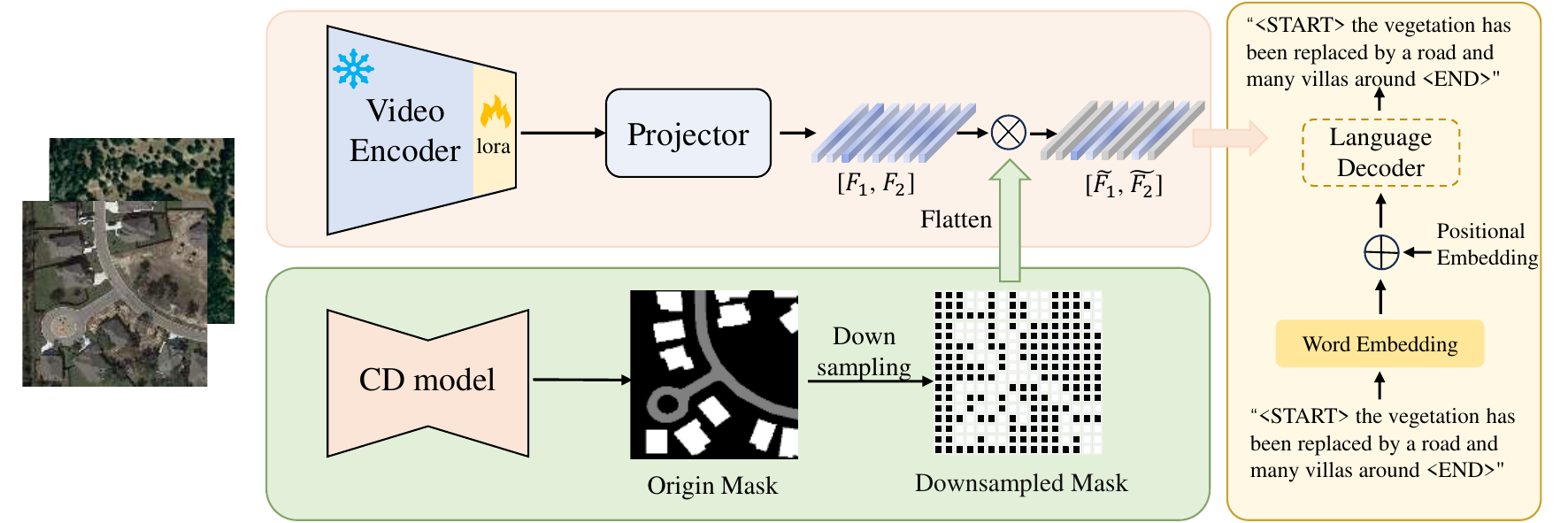}
   \caption{The detailed illustration of MV-CC. The pre-trained video encoder is applied with LoRA \cite{loralowrankadaptationlarge} fine-tuning to extract temporal features, followed by a simple projector to generate corresponding tokens. The mask obtained from the CD model is downsampled and multiplied with the tokens to produce new tokens (tokens in gray represent those that have been deactivated.). These tokens are then fed into a language decoder.}
   \label{flowchart}

\end{figure*}
In this paper, we propose a video encoder-decoder two-stage architecture, which differs from the traditional composition of feature extractors, feature fusion, and caption decoders. The whole architecture can be formulated as:

\begin{equation}
\begin{aligned}
[F_1,F_2] = \phi_{video\_encoder}([X_1,X_2])
\end{aligned}
\end{equation}

\vspace{-1em}
\begin{equation}
\begin{aligned}
E_{text} = \phi_{decoder}([F_1, F_2]),
\label{eq1}
\end{aligned}
\end{equation}
where $X_1\in \mathbb{R}^{H \times W \times 3}$ and $X_2\in \mathbb{R}^{H \times W \times 3}$ indicate a pair of bi-temporal images, $\phi_{video\_encoder}$ indicates the video encoder, and $\phi_{decoder}$ indicates the language decoder. $F_1\in \mathbb{R}^{hw \times c}$ and $F_2\in \mathbb{R}^{hw \times c}$ denote the image tokens, where $h=H/s$, $w=W/s$, and $s$ denote the stride size (e.g., $s$=16 for ViT/16). $E_{text}$ denotes the generated caption.



In this new paradigm, paired bi-temporal images are fed as two video frames to the video encoder, where the video encoder conducts spatiotemporal feature extraction and fusion through its built-in 3D CNN or ViT. At the end of this process, an optional projector maps the encoder's image features into the text embedding space. Finally, these features are fed into the language decoder to generate change captions. Compared to the traditional three-stage paradigm, our two-stage paradigm can utilize the video model's inherent temporal extraction capability in a more concise and graceful manner. We describe the video-based CC paradigm and its instantiation in detail in the following.


\subsection{Video-based Encoder}




When processing bi-temporal data, compared to plain Siamese networks, the video-based model has a stronger ability to capture temporal signals, while avoiding the manual design of the fusion module after the Siamese network. For the sake of illustration, here, we take InternVideo2's stage3-1B model \cite{Internvideo} as the video model of MV-CC, which is pre-trained through unmasked video token reconstruction, multimodal contrastive learning, and next token prediction.

To overcome the gap between video data and bi-temporal images, we consider the bi-temporal remote sensing images as two frames of video and concatenate them as the input to the video model. In InternVideo2, spatiotemporal features between bi-temporal images are first captured by 3D convolution, and then these preliminary features are fed into the ViT model (blocks). The ViT model focuses on the subtle and complex spatial information of the image, and due to its long-range dependency, it also deeply focuses on the temporal relationship between bi-temporal images.

Considering that the model is pre-trained on natural videos, to make the frozen video model learn and understand the domain information on remote sensing images, we fine-tune linear layers of the last block of the ViT with Low-Rank Adaptation (LoRA)\cite{loralowrankadaptationlarge}. LoRA is a technique that allows us to fine-tune the model by updating only a small fraction of the parameters, which helps in adapting the model to new tasks without significantly increasing the computational cost. Specifically, we only use LoRA to fine-tune the last Transformer block to avoid high graphics memory usage:

\begin{equation}
\begin{aligned}
\label{lora}
W_{fs1}^{\text{new}} &= W_{fs1} + (W_{lora\_A} \cdot W_{lora\_B}), \\
W_{fs2}^{\text{new}} &= W_{fs2} + (W_{lora\_A} \cdot W_{lora\_B}),
\end{aligned}
\end{equation}
where $W_{fs1}^{\text{new}}$ and $W_{fs2}^{\text{new}}$ denote the updated weight matrices after applying LoRA. $W_{lora\_A}$ and $W_{lora\_B}$ denote the low-rank matrices used in LoRA, designed with a rank of 4.

Furthermore, to transforms the image tokens from the video encoder into the text embedding space and matches the dimension size of the language decoder, we design a learnable projector, which is defined as: 

\begin{equation}
\label{linear}
[\widetilde{F_1},\widetilde{F_2}] = \text{Projector}([F_1,F_2]).
\end{equation}

Here, we instantiate the projector as a simple linear layer.

\subsection{Mask enhanced branch}



The zero-shot results of the large video model, as shown in Fig. \ref{interestiong}, indicate that the video encoder effectively focuses on the changes of interest (roads, buildings, etc.). However, some changes of non-interest such as vegetation and lighting are also described, which is due to the fact that the video model is trained on extensive natural videos and it is difficult to provide an accurate description for specialized tasks. Therefore, guiding the model to focus on the correct regions is crucial, specifically the ViT architecture, i.e., how to filter out irrelevant image tokens. Inspired by the CD task, which can generate a segmentation mask containing change and unchanged categories, we propose an explicit guidance method based on the prediction mask from the CD task.


For a single frame input, the sequence length of the ViT output tokens is $h\times w$ (ignoring \texttt{[CLS]} token), which corresponds to a $s\times$ downsampling of the original image input. Since the sequence length is different from the scale of the CD mask, it is first necessary to execute a downsampling operation for the CD mask:
\begin{equation}
\label{downsample}
Mask_{\downarrow} = \text{Downsample}_{\text{nearest}}(Mask, h, w),
\end{equation}
where $Mask \in \{0,1\}^{H \times W}$ represents the prediction mask of CD model, and $Mask_{\downarrow} \in \{0,1\}^{h \times w}$ represents the downsampled mask, which is then flattened and multiplied with image tokens to filter out irrelevant tokens:

\begin{equation}
Mask^{\prime}_{\downarrow} = \text{Flatten}(Mask_{\downarrow}),
\end{equation}
Eq. \ref{eq1} is improved as:
\begin{equation}
\begin{aligned}
E_{text} = \phi_{decoder}([\widetilde{F_1} \odot Mask^{\prime}_{\downarrow}, \widetilde{F_2} \odot Mask^{\prime}_{\downarrow}]),
\label{mask}
\end{aligned}
\end{equation}
where $\odot$ denotes the Hadamard product, and $Mask^{\prime}_{\downarrow}$ broadcasts in the channel dimension of $F_1$ and $F_2$. Based on this explicit guidance method, the image tokens corresponding to unchanged regions in the CD prediction mask are discarded, and only the effective tokens of the changed regions are fed into the subsequent language decoder to generate accurate, task-specific captions.


As for \textbf{how to obtain the CD mask}, with the development of the CD task, a series of methods have been proposed, including unsupervised, semi-supervised, and fully supervised methods, all of which can obtain the aforementioned mask. Therefore, the Mask-enhanced branch has strong adaptability and generalization ability and can be used with our model in cases with varying amounts of CD annotations. Significantly, the requirement for low-resolution masks further reduces the additional workload and the precision requirements of the CD model.

\subsection{Change Caption Decoder}
For the language decoder module, we select the standard Transformer decoder\cite{attentionneed,bertpretrainingdeepbidirectional}, a typical model frequently used in the field of natural language processing. This decoder consists of multiple Transformer layers, with each layer composed of three sub-networks: a masked multi-head attention module, a cross-attention module, and a feed-forward network. Each sub-network is accompanied by residual connections and layer normalization. Additionally, to preserve the information in the word embeddings, residual connections are also applied across the entire decoder layer. Finally, the output is generated through a linear layer followed by a Softmax activation function.

Before feeding the text information into the decoder, the text tokens need to be processed. First, the text tokens $t$ need to be mapped to word embeddings through a mapping function 
$f_{emb} := \mathbb{R}^n  \rightarrow \mathbb{R}^{n \times d_{emb}}$ to obtain the word embeddings. Since attention is insensitive to positional information, additional positional embeddings $E_{pos}$ are required. Therefore, the original input of the decoder can be represented as:
\begin{equation}
    E_{text}^0 = f_{emb}(t) + E_{pos}.
\end{equation}

After this, the masked multi-head attention is used to aggregate text information. It is important to note that the attention mechanism may lead to leakage of future text, which could harm the model's performance. To prevent this, a lower triangular mask with the same dimensions as the weight matrix is constructed to help the model focus on positions in the output sequence that have already been generated, avoiding information leakage.

The difference between the generated change captions and the ground truth can guide the training of the entire model, with the loss function used being cross-entropy loss:
\begin{equation}
    L = -\sum_{t=1}^Tlog(\sum_{k=1}^Ky_{t,k}p_{t,k}),
\end{equation}
where $T$ represents the length of the captions, $K$ represents the size of vocabulary. $y_t = [y_t^{(1)}, y_t^{(2)},\ldots, t_t^{(K)}]$ represents the vector representation of the word at position $t$ in ground truth, the value of $y_{t,k}$ is 1 if word k is at position $t$, otherwise 0. And $p_t = [p_t^{(1)}, p_t^{(2)},\ldots, p_t^{(K)}$ is the probability vector for predicting the word at position $t$ in the generated sentence.

\section{Experimental Results and Analyses} \label{sec:exp}

\begin{table*}[t]
\centering{
\caption{Compare with other state-of-the-art methods.}
\label{t1}
\begin{tabular}{cc|c c c c c c c| c }
	\toprule
	Fusion module Type & Method & BLEU-1 & BLEU-2 & BLEU-3 & BLEU-4 & METEOR & ROUGE$_L$ & CIDEr-D  & Trainable Param. \\
	\midrule
	\multirow{4}{*}{\shortstack{CNN \\based}} & {Capt-Rep-Diff}\cite{robustchangecaptioning} & 72.90 & 61.98 & 53.62 & 47.41 & 34.47 & 65.64 & 110.57  & 73.21M \\
	 & {Capt-Att}\cite{robustchangecaptioning} & 77.64 & 67.40 & 59.24 & 53.15 & 36.58 & 69.73 & 121.22  & 73.60M \\
	 & {Capt-Dual-Att}\cite{robustchangecaptioning} & 79.51 & 70.57 & 63.23 & 57.46 & 36.56 & 70.69 & 124.42 &  75.58M \\
	 & {DUDA }\cite{robustchangecaptioning} & 81.44 & 72.22 & 64.24 & 57.79 & 37.15 & 71.04 & 124.32 &  80.31M \\
  \cline{1-2}
	 \multirow{6}{*}{\shortstack{Transformer\\based}} & {MCCFormer-S}\cite{describinglocalizingmultiplechanges} & 79.90 & 70.26 & 62.68 & 56.68 & 36.17 & 69.46 & 120.39 & 162.55M \\
	 & {MCCFormer-D}\cite{describinglocalizingmultiplechanges} & 80.42 & 70.87 & 62.86 & 56.38 & 37.29 & 70.32 & 124.44  & 162.55M\\
	 & {RSICCFormer}\cite{RSICCFormer} & {84.72} & {76.27} & {68.87} & {62.77} & {39.61} & {74.12} & {134.12} & 172.80M\\
      & {PSNet}\cite{PSNet} & 83.86 & 75.13 & 67.89 & 62.11 & 38.80 & 73.60 & 132.62 & 319.76M\\
      & {PromptCC}\cite{PromptCC} & 83.66 & 75.73 & 69.10 & 63.54 & 38.82 & 73.72 & 136.44 & 408.58M\\
       &{Chg2Cap}\cite{chg2cap} & 86.14 &78.08 &70.66& 64.39& 40.03& 75.12 &136.61& 285.5M \\
       \cline{1-2}
 SSM based & {RSCaMa}\cite{rscamaremotesensingimage} & 85.79 & 77.99 & 71.04 & 65.24 & 39.91 & 75.24 & 136.56  & {176.90M} \\
 	\midrule
None & {MV-CC} & \textbf{86.37} & \textbf{79.01} & \textbf{72.03} & \textbf{66.22} & \textbf{40.20} & \textbf{75.73} & \textbf{138.28} & {119.48M} \\
	\bottomrule
\end{tabular}}
\end{table*}

\begin{table*}[h]
\centering{
\caption{Compare with other methods with CD mask.}
\label{t2}
\begin{tabular}{c|c c c c c c c}
	\toprule
	 Method & BLEU-1 & BLEU-2 & BLEU-3 & BLEU-4 & METEOR & ROUGE$_L$ & CIDEr-D \\
	\midrule
	 { Detection Assited }\cite{Detection}&87.26 &79.16 &71.74& 65.56 & 40.65 & 75.97 & 139.99\\
	 {Semantic-CC}\cite{Zhu2024SemanticCCBR} &\textbf{88.07} & \textbf{79.68} & 71.47 & 64.51 &  40.58 & \textbf{77.76} & 138.51\\
   {KCFI} \cite{yang2024enhancingperceptionkeychanges} & 86.34& 77.31& 70.89& 65.30& 39.42& 75.47& 138.25 \\
    {ChangeMinds}\cite{changemindsmultitaskframeworkdetecting} & 86.39 & 78.34 & 71.35& 65.60& \textbf{40.86}& 75.85& \textbf{140.32} \\
   {MV-CC w/o testset} & 85.18& 77.43& 70.73& 65.28& 39.83 & 74.99& 137.50\\
 {MV-CC w/ testset} & 86.37 & 79.01 & \textbf{72.03} & \textbf{66.22} & 40.20 & 75.73 & 138.28 \\
	\bottomrule
\end{tabular}}
\end{table*}
\subsection{Datasets}
We conducted our experiments using the LEVIR-MCI dataset\cite{dataset}, which contains 10,077 pairs of bitemporal images. Each image has a resolution of $256\times256$ pixels with a high spatial resolution of 0.5 meters per pixel. Each image pair is annotated with corresponding CD masks and descriptive captions. This dataset is an extension of the LEVIR-CC\cite{LevirCC} dataset, with additional CD masks that highlight changes in roads and buildings. The division of the training, validation, and test sets is consistent with the setup in LEVIR-MCI.

\subsection{Experimental Setup}
\begin{enumerate}
    \item \textbf{Evaluation Metrics.}
To evaluate the similarity and quality between the generated text descriptions and reference descriptions, we selected commonly used metrics in CC task. Specifically, we employed BLEU-1, BLEU-2, BLEU-3, BLEU-4, METEOR, ROUGE, and CIDEr in our experiments. These metrics comprehensively assess the accuracy, fluency, semantic relevance, and consistency with the visual content of the generated descriptions.
    \item \textbf{Implementation Details.}
Our proposed MV-CC model is implemented using the PyTorch framework and is trained and evaluated on a single NVIDIA GeForce RTX 4090 GPU. During training, the Adam optimizer is employed, with the learning rate, number of training epochs, and batch size kept consistent with Chg2Cap method \cite{chg2cap}. After each epoch, the model is evaluated on the validation set, and the model with the highest BLEU-4 score is selected for evaluation on the test set.
\end{enumerate}
\subsection{Comparison with the State-of-the-Art}
Existing studies have employed Siamese networks to extract features, focusing on the ``neck'' design of the model to integrate features, which includes both CNN architectures and Transformer architectures. Recent work, such as RSCaMa\cite{rscamaremotesensingimage}, proposed to use the Structured Space Model (SSM) for feature fusion. Ours MV-CC utilizes a video model to accomplish both feature extraction and feature fusion simultaneously.

In our comparison of the MV-CC model with state-of-the-art (SOTA) methods, we have demonstrated superior performance across the most critical parameters. As shown in Table \ref{t1}, the MV-CC model exhibits a distinct advantage in all performance metrics. Specifically, when compared with the SOTA method (RSCaMa), we achieved a 1\% improvement in BLEU-4 scores and reduced the number of trainable parameters by 32\%. Among them, the trainable parameters of the video encoder are only 3M.

We compare the method guided by CD labels as shown in Table \ref{t2} with the guidance method of MC-CC. When both training and testing set labels are available, we achieve a performance improvement of 0.4\% in BLEU-4. Although our performance is lower than some methods when the test set labels are unknown, the labels we require are only at 1/16th the resolution of the original image, while several other models require the resolution of the original image size ($256\times 256$). Therefore, the CD guidance method we propose greatly reduces the requirement for labels and can achieve significant performance improvements at a very low annotation cost. Furthermore, MV-CC relies on established CD models, and with the advancement of CD methodologies, there is potential for continuous improvement in our proposed method.


\subsection{Ablation Studies}

\begin{table*}[t]
\centering{
\caption{Experiments verifing the effectiveness of different strategies on the LEVIR-MCI dataset. LoRA: fine-tuning using lora. Mask: mask enhanced branc. proportion: the proportion of labels used in the training set.}
\label{t3}
\begin{tabular}{c|llc c c c c c c c}
    
	\toprule
	 Method  & LoRA  & Mask & Proption & BLEU-1 & BLEU-2 & BLEU-3 & BLEU-4 & METEOR & ROUGE$_L$ & CIDEr-D \\
	\midrule
	  & $\times$& $\times$& 0\%&84.20&75.42&68.07& 62.01&38.18 & 73.19& 130.40\\
	  MV-CC& $\checkmark$& $\times$&0\%& 83.80& 75.13& 68.09& 62.44& 38.31 & 73.29& 132.49\\
   & $\checkmark$& $\checkmark$& 5\%& 85.37& 76.90& 69.85& 64.10& 39.96 & 74.92& 137.96\\
   
 & $\checkmark$& $\checkmark$& 100\%& 85.18& 77.43& 70.73& 65.28& 39.83 & 74.99&137.50\\ \hline
\end{tabular}}
\end{table*}

\begin{figure*}[t]
  \centering
   \includegraphics[width=0.95\linewidth]{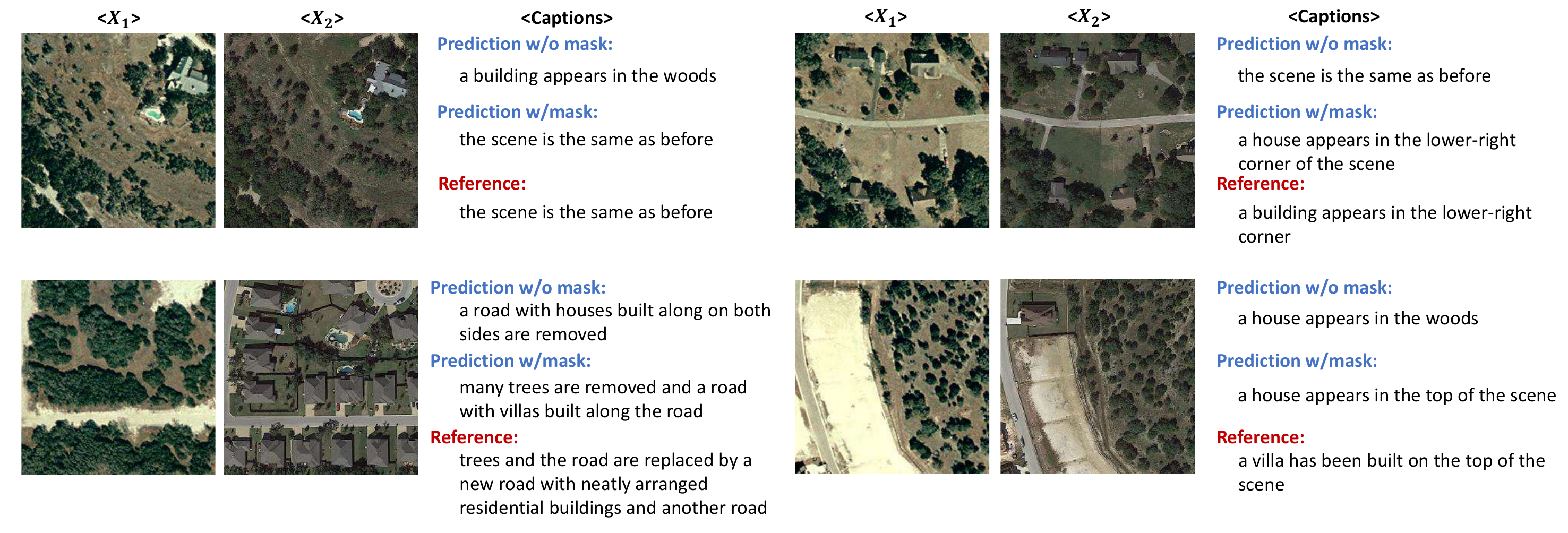}
   \caption{Change captioning examples generated by MV-CC, which demonstrate the difference in descriptions with and without mask guidance in the LEVIR-MCI dataset.}
   \label{changeCaption}
   \vspace{-4mm}
\end{figure*}

\textbf{LoRA Fine-tuning:} Due to the pre-training of the video encoder on a large amount of visible light videos, we opt for LoRA fine-tuning of the model to adapt it to the task of remote sensing imagery. Specifically, we introduce LoRA layers with a rank of 4 to the three linear layers of the last block of the model, as shown in Table \ref{t3}, achieving a 0.4\% improvement in BLEU-4.

\textbf{Mask Proportion:} We tested the semi-supervised method with 5\% of the training set labels to obtain masks and the fully supervised method with 100\% of the training set labels to obtain masks. As shown in Table \ref{t3}, compared to the situation without mask guidance, both scenarios achieved improvements of 1.7\% and 2.8\% respectively. The experimental results prove that the mask-guided method effectively increases the model's focus on the changing areas, and this approach is expected to be combined with more advanced models in the future to enhance their understanding of changing areas. 
Fig. \ref{changeCaption} shows that the model guided by the mask can focus on the true areas of change. In the second pair of bi-temporal images, there are only minor changes in the lower right corner; nevertheless, the model with the mask can still accurately identify and describe this change. Additionally, the model with mask guidance effectively locates the regions of change, providing more accurate captions, as confirmed by the fourth pair of bi-temporal images.

\section{Conclusion} \label{sec:con}
This paper introduces MV-CC, a new paradigm for addressing remote sensing change captioning issues. We explored the application of the video models in remote sensing change caption and first proposed that a pure video model can simultaneously extract temporal and spatial features from bi-temporal images without the need for additional fusion modules. Furthermore, considering the maturity of the current CD field, we leveraged existing CD models to guide the CC model, enabling it to accurately focus on the areas that have undergone genuine changes. Notably, our model is naturally adaptable to potential future multi-temporal change captioning tasks, and its performance has broad room for improvement as CD methods are refined. We believe that the exploration of video-based CC models goes beyond this, and more effective utilization of features from video models or guiding the model's attention in other ways are meaningful research directions for the future.

\bibliographystyle{IEEEtran}

\bibliography{main}

\vfill

\end{document}